\documentclass[11pt]{article}

\usepackage[preprint]{acl}

\usepackage{times}
\usepackage{float}
\usepackage{latexsym}
\usepackage{amsmath}
\usepackage{amssymb}
\usepackage{multirow}
\usepackage{subcaption}
\usepackage{booktabs} 
\usepackage{makecell}
\usepackage{tcolorbox}
\usepackage{xcolor}
\usepackage{tcolorbox}
\tcbuselibrary{skins,breakable}
\usepackage{siunitx}
\usepackage{adjustbox}
\usepackage[table]{xcolor}

\newcommand{\wrong}[1]{\textcolor{red}{#1}}
\newcommand{\correct}[1]{\textcolor{green!50!black}{#1}}
\definecolor{mycolor}{RGB}{210,240,210}
\newcommand{\cc}{\cellcolor{mycolor}}

\usepackage[T1]{fontenc}
\usepackage[utf8]{inputenc}

\usepackage{microtype}

\usepackage{inconsolata}

\usepackage{graphicx}

\title{Less is More: \\ Improving LLM Reasoning with Minimal Test-Time Intervention}

\author{
 \textbf{Zhen Yang\thanks{This work was conducted during the author's internship at Kling Team.}\textsuperscript{1}},
 \textbf{Mingyang Zhang\textsuperscript{5}},
 \textbf{Feng Chen\textsuperscript{3}},
 \textbf{Ganggui Ding\textsuperscript{4}},
\\
 \textbf{Liang Hou\textsuperscript{2}},
 \textbf{Xin Tao\textsuperscript{2}},
 \textbf{Ying-Cong Chen\thanks{Corresponding author.}\textsuperscript{1,6}}
\\
\\
 \textsuperscript{1}HKUST(GZ),
 \textsuperscript{2}Kuaishou Technology,
 \textsuperscript{3}AIML,
 \textsuperscript{4}ZJU,
 \textsuperscript{5}Ant Group,
 \textsuperscript{6}HKUST
\\
 \small{
   \href{zheny.cs@gmail.com}{zheny.cs@gmail.com}, \href{yingcongchen@ust.hk}{yingcongchen@ust.hk}
 }
}

\begin{document}
\maketitle

\begin{abstract}
Recent progress in large language models (LLMs) has focused on \emph{test-time scaling} to improve reasoning via increased inference computation, but often at the cost of efficiency. We revisit test-time behavior and uncover a simple yet underexplored phenomenon: reasoning uncertainty is highly localized—only a small subset of high-entropy tokens dominantly affects output correctness. Motivated by this, we propose \textbf{Minimal Test-Time Intervention (MTI)}, a training-free framework that enhances reasoning accuracy and stability with minimal overhead. MTI includes: (i) \emph{Selective CFG intervention}, applying classifier-free guidance only at uncertain positions; and (ii) \emph{Lightweight negative-prompt guidance}, reusing the main model’s KV cache to approximate unconditional decoding efficiently. MTI yields consistent gains across general, coding, and STEM tasks—e.g., +9.28\% average improvement on six benchmarks for DeepSeek-R1-7B and +11.25\% on AIME2024 using Ling-mini-2.0—while remaining highly efficient.
% The anonymous vLLM version code can be found \href{https://anonymous.4open.science/r/MTI_anonymous}{here}.
The code can be found \href{https://github.com/EnVision-Research/MTI}{here}.
\end{abstract}

\section{Introduction}

Large language models (LLMs)~\citep{gpt4,gemini,llama,deepseek,qwen3,starcoder,yi} have advanced rapidly, with ever-increasing parameter counts, larger pretraining corpora, and improved optimization techniques leading to steady gains across benchmarks. 
Beyond such \emph{training-time scaling}, recent research has increasingly focused on \emph{test-time scaling}—allocating more computation during inference through deeper reasoning or broader search. 
Methods such as chain-of-thought prompting~\citep{wei2022chain}, self-consistency~\citep{wang2022self}, and frameworks like Reflexion~\citep{shinn2023reflexion} or Tree of Thoughts~\citep{yao2023tree} demonstrate that increasing inference-time computation can enhance reasoning quality and output accuracy. 
However, these methods achieve better performance \emph{by trading inference efficiency for computation}, as they require multiple reasoning trajectories, extensive sampling, or long multi-step deliberation, making them computationally expensive and sometimes impractical in real-world deployments. 

\begin{figure}[t]
  \centering
  \includegraphics[width=\linewidth]{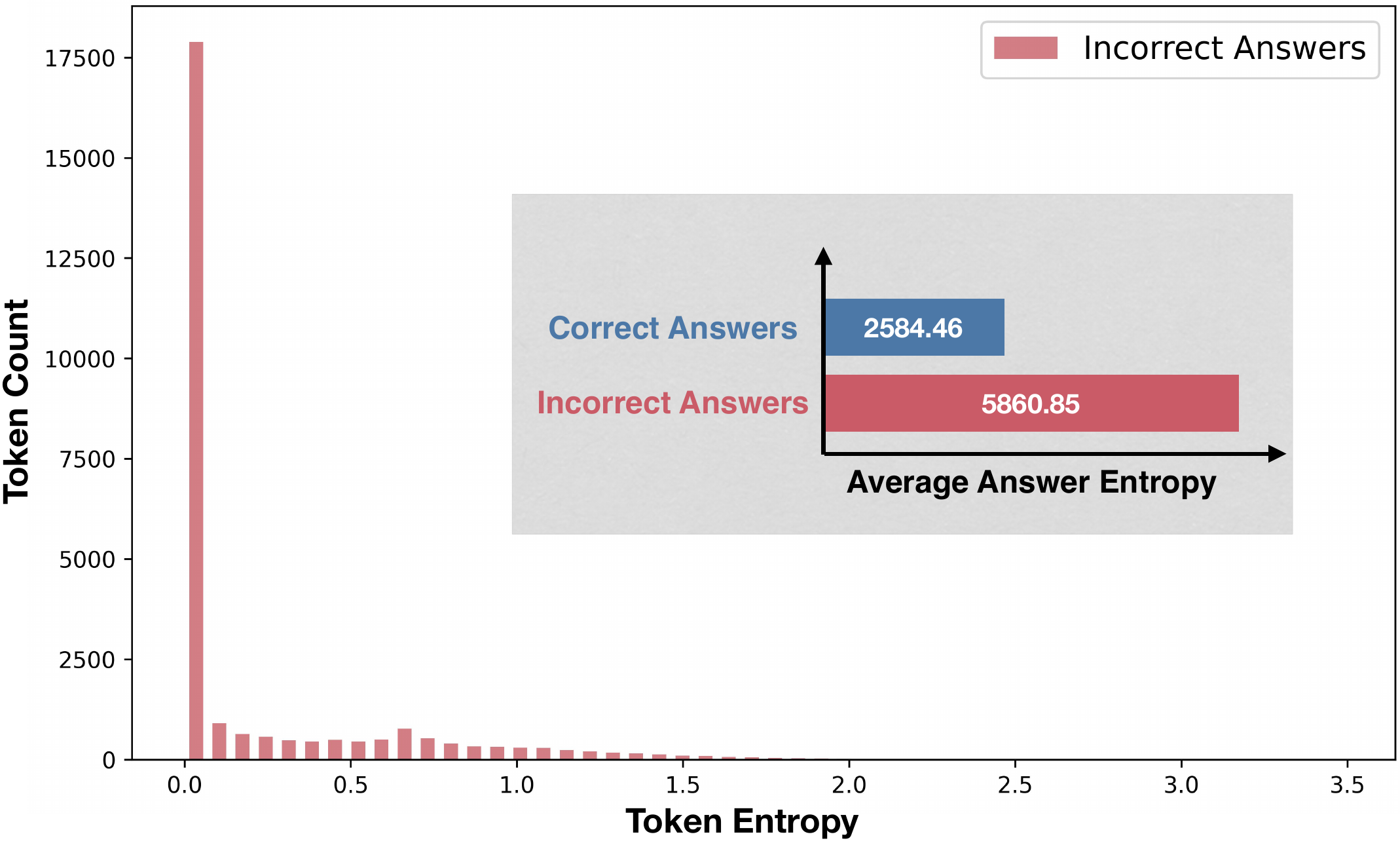}
  \caption {Incorrect answers exhibit higher average answer entropy than correct ones, primarily due to high-entropy tokens.
  % (results for Qwen3-8B-Reasoning on AIME2024). 
  }
  \label{fig:observation}
\end{figure}

In this work, we challenge the prevailing assumption that improving test-time performance of large language models must inherently come at the cost of increased inference computation. 
To examine this, we conduct an empirical analysis on the AIME2024 benchmark, comparing reasoning trajectories that yield correct versus incorrect answers. 
Interestingly, as shown in the inset panel in the upper-right corner of Fig.~\ref{fig:observation}, correct responses exhibit significantly lower average answer entropy than incorrect ones, indicating that reasoning failures are associated with higher uncertainty. 
Furthermore, as illustrated in the main panel in Fig.~\ref{fig:observation}, only a small fraction of tokens within each response have notably high entropy, yet these few tokens contribute a disproportionate share of the average answer entropy.
This observation suggests that reasoning errors are not uniformly distributed across the sequence but are instead concentrated in a few high-entropy ``critical steps'', where local uncertainty can propagate and destabilize subsequent predictions. 
% We hypothesize that by selectively stabilizing these high-entropy regions
% —e.g., through targeted guidance that reduces token-level entropy—
% , one can suppress error amplification and improve overall reasoning accuracy and consistency, all while intervening on only a minimal subset of tokens.
We hypothesize that by selectively stabilizing these high-entropy regions, one can suppress error amplification and improve overall reasoning accuracy and consistency, all while intervening on only a minimal subset of tokens.

Building on this observation, we propose \textbf{Minimal Test-Time Intervention (MTI)}, a training-free approach that improves reasoning performance without incurring substantial inference cost.   
Specifically, MTI first detects high-entropy tokens by measuring the dispersion of the model’s predictive distribution during generation. 
High entropy indicates uncertainty in token prediction, and thus a greater likelihood that local instability may propagate through the autoregressive chain. 
To reduce this uncertainty, MTI draws inspiration from \emph{classifier-free guidance (CFG)}~\citep{cfg_llm}, which combines conditional and unconditional predictions to steer model outputs toward more task-aligned regions of the probability space. 
However, directly applying standard CFG would require maintaining a separate unconditional branch, resulting in computational and memory overhead.

To address this, MTI introduces a lightweight alternative: before generating a high-entropy token, the model temporarily augments its input with the short phrase ``\texttt{OUTPUT ERROR}'', which implicitly constructs an on-the-fly unconditional contrastive branch for CFG-style guidance. 
This branch reuses the main model’s KV cache, adding only two auxiliary tokens per high-entropy position for the CFG computation. 
Since high-entropy tokens constitute only a small fraction of the overall sequence, the additional overhead is negligible. 
In practice, this design effectively shifts the model’s predictive distribution toward the correct region of the probability space, enhancing reasoning stability and accuracy with negligible test-time computation.

In general, our main contributions are as follows:

\begin{itemize}
  \item We find that high-entropy tokens are the main sources of error in LLM reasoning. Detecting them via token entropy and intervening selectively improves reasoning performance.
  \item We introduce a training-free test-time intervention based on CFG that selectively applies guidance to high-entropy tokens, yielding better answers with minimal overhead. We further implement negative-prompt injection via KV-cache reuse, which shrinks the unconditional branch’s KV-cache allocation and more faithfully approximates the unconditional distribution used by CFG.
  \item We achieve robust performance across general, coding, and STEM benchmarks, showing consistent improvements over direct inference across the latest LLMs of varying scales. Specifically, our method yields a 9.28\% average gain on six comprehensive tasks with DeepSeek-R1-7B and a +11.25\% improvement on the AIME2024 benchmark with Ling-mini-2.0, demonstrating its broad effectiveness across different tasks.
\end{itemize}

\begin{figure*}[!t]
  \centering
  \includegraphics[width=\linewidth]{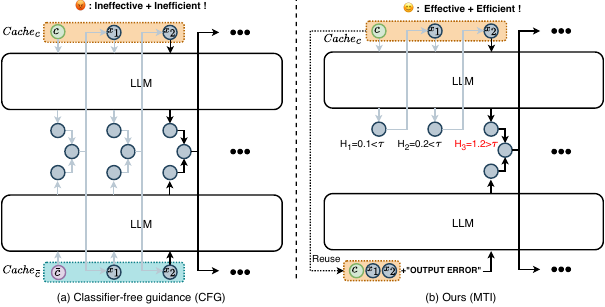}
    \caption{
    % Comparison between CFG and MTI. MTI applies CFG only to high-entropy tokens and reuses the conditional KV cache with an injected negative prompt, avoiding the unconditional cache and enabling both higher efficiency and a better approximation of the unconditional distribution, thereby suppressing erroneous content generation.
    Comparison between CFG and MTI. MTI applies CFG exclusively to high-entropy tokens and reuses the conditional KV cache via negative prompt injection. 
    % This approach not only improves computational efficiency and memory overhead by bypassing the unconditional cache but also achieves a superior approximation of the unconditional distribution, thereby effectively suppressing erroneous content generation.
    }
  \label{fig:method}
\end{figure*}

\section{Related Works}

\noindent\textbf{Test-time optimizations for LLMs. }Recent research focuses on test-time or decoding-time optimizations to enhance LLM reasoning, robustness, or efficiency without retraining~\citep{chen2025parallel}. One line of research improves reasoning capabilities through prompt engineering and advanced sampling strategies, such as Chain-of-Thought (CoT)~\citep{wei2022chain}, self-consistency~\citep{wang2022self}. Building on these, search and evaluation frameworks have been proposed, including Reflexion~\citep{shinn2023reflexion}, Tree of Thoughts~\citep{yao2023tree}, DeepConf~\citep{fu2025deep}, and EGB~\citep{gatebranch}. Despite their effectiveness in enhancing semantic robustness, these methods often incur significant sampling and memory overhead~\citep{cfg_llm}. Furthermore, hallucination mitigation strategies such as Contrastive Decoding~\citep{CD,UCD,xu2026thinking} introduce additional complexity by incorporating an auxiliary small model. Another direction targets computational efficiency via speculative decoding~\citep{SD, R-Stitch, yan2024decoding, zhang2023draft} or memory optimizations like FlashAttention~\citep{dao2022flashattention}. 
Unlike these computationally intensive approaches, our method selectively intervenes only on high-entropy tokens, achieving significant performance improvements with negligible overhead. Furthermore, our framework is highly flexible, serving as a plug-and-play decoding strategy that can be seamlessly combined with existing reasoning techniques.
We further discuss its connection with other decoding strategies in Sec.~\ref{sec:ds_related_work}.

\noindent\textbf{Classifier-free guidance in LLMs. }Originally developed for diffusion models, Classifier-Free Guidance (CFG)~\citep{cfg} balances fidelity and diversity by interpolating between conditional and unconditional predictions. This mechanism can also incorporate negative prompts to steer generation away from undesirable attributes, such as content errors~\citep{dynamicnp} or artifacts~\citep{npi, oir}. However, adapting CFG to LLMs~\citep{cfg_llm} faces two primary hurdles: the lack of a well-defined unconditional space—rendering empty negative prompts poorly calibrated—and the huge computational and memory overhead. We address these by applying CFG selectively to high-entropy tokens and introducing negative-prompt injection via KV-cache reuse, effectively eliminating redundant cache allocation and computational inefficiency while improving unconditional space approximation to suppress erroneous content generation.

\noindent\textbf{Entropy-based token selection in LLMs. }Recent studies emphasize that LLM tokens exhibit varying uncertainty, with token-level entropy serving as a reliable metric for identifying critical decision boundaries. While prior work leverages low-entropy tokens for redundancy pruning~\citep{li2025compressing, xia2025tokenskip} or focuses on high-entropy tokens for optimization~\citep{beyond8020, lee2025well, qian2025demystifying}, these methods typically require retraining or heuristic-based pruning. In contrast, our approach is entirely training-free and operates at inference-time. Specifically, we leverage token entropy to determine where to apply CFG—only to tokens with high entropy.

\section{Method}

\subsection{Preliminary}
\label{sec:Preliminary}

We first introduce the notation. Let $P(\cdot)$ denote the token distribution predicted by the LLM. Given a context $c$, the conditional probability of generating a token $x$ is denoted by $P(x \mid c)$. In the setting of classifier-free guidance (CFG), $\bar{c}$ represents a negative prompt in the unconditional branch. The CFG-adjusted distribution is denoted by $\hat{P}(\cdot)$, where $\omega$ controls the strength of guidance. We use $l_t$ to denote the logits predicted at step $t$, $x_t$ denotes the token generated at step $t$, and $\tau$ denotes the entropy threshold. Simple examples of CFG and negative prompts can be found in Sec.~\ref{sec:Negative_prompts_in_CFG_for_LLMs}.

\noindent\textbf{Classifier-Free Guidance in LLMs.}
Originally proposed for diffusion models, classifier-free guidance~\citep{cfg} improves conditional generation by interpolating between conditional and unconditional predictions. In diffusion models, an additional null condition is introduced during training to learn the global data distribution. In contrast, for LLMs, CFG is applied purely at inference time without requiring additional training objectives~\citep{cfg_llm}. The guided prediction is formulated in the log-probability space as:
\begin{equation}
\begin{array}{l}
\log \hat{P}\left(x_t \mid c,\bar{c},x_{<t}\right) \\
  = (1-\omega) \cdot \log P\left(x_t \mid \bar{c}, x_{<t}\right) \\
  \quad {}
  + \omega \cdot \log P\left(x_t \mid c, x_{<t}\right),
\end{array}
\label{eq:cfg}
\end{equation}
where $x_{<t}$ denotes the sequence of previously generated tokens up to step $t-1$. The negative prompt $\bar{c}$ encourages the model to generate outputs that deviate from the unconditional distribution.

\noindent\textbf{Token and answer entropy.}
\label{sec:token_level_entropy}
We adopt Shannon entropy to quantify the model’s uncertainty over predicted logits. Lower entropy indicates higher confidence, while higher entropy reflects greater uncertainty. The token entropy at step $t$ is defined as:
\begin{equation}
H_t = -\sum_{i=1}^{V} p_i \log p_i,
\label{eq:entropy}
\end{equation}
where $\left[p_1, \ldots, p_V\right]_t = \operatorname{softmax}(l_t)$, $V$ denotes the vocabulary size, and $p_i$ is the probability assigned to the $i$-th token.

The answer-level entropy for the $n$-th question is defined as the sum of token entropies along its full reasoning trajectory:
\begin{equation}
A_n = \sum_{t=1}^{T_n} H_t,
\end{equation}
where $T_n$ is the length of the generated trajectory. The average answer entropy $A_{avg}$ over $N$ data instances is given by:
\begin{equation}
A_{avg} = \frac{1}{N} \sum_{n=1}^{N} A_n.
\label{eq:answer_entropy}
\end{equation}

\noindent\textbf{KV Cache in LLM-CFG.}
At generation step $t$, the key-value (KV) cache avoids recomputation of full-prefix attention by reusing previously stored keys and values. Under LLM-CFG, two separate KV caches are maintained—one for conditional prediction and one for unconditional prediction—to enable efficient guided inference. Specifically, the caches used to compute the outputs for the two branches at step t are:
\begin{equation}
\begin{split}
\mathrm{Cache}_{c} &= \left[c, x_{1}, \ldots, x_{t-1}\right], \\
\mathrm{Cache}_{\bar{c}} &= \left[\bar{c}, x_{1}, \ldots, x_{t-1}\right].
\end{split}
\label{eq:kvcache}
\end{equation}
For simplicity, we do not distinguish between keys and values; for example, $c$ denotes the KV pairs cached after processing the prompt $c$. Fig.~\ref{fig:method} follows the same notation.

\subsection{Selective CFG intervention}
LLMs often display chain instability in multi-step reasoning: uncertainty at a few steps amplifies and derails the entire answer. To locate where this instability originates, 
the inset panel in the upper-right corner of Fig.~\ref{fig:observation} shows that questions answered incorrectly have markedly higher average answer entropy than those answered correctly. The main panel further indicates that the gap is driven by high-entropy tokens in erroneous responses. These findings suggest that overall failure is concentrated at a small set of critical high-entropy nodes and that stabilizing these nodes can yield gains. 

Based on this observation, Fig.~\ref{fig:method}b proposes a selective CFG intervention strategy. During decoding, we monitor token entropy and intervene only when it exceeds a threshold $\tau$: tokens with entropy $\le \tau$ proceed normally, whereas tokens with entropy $> \tau$ receive CFG to prevent error propagation. Compared with the vanilla CFG baseline in Fig.~\ref{fig:method}a, which applies guidance uniformly at every step, our approach activates selectively at unstable nodes, concentrating guidance where it matters most and improving both stability and efficiency.

\subsection{Lightweight negative-prompt guidance}
However, as illustrated in Fig.~\ref{fig:method}a, although selective CFG intervention reduces the frequency of application, it still requires maintaining two separate sets of KV caches because the contexts of the conditional and unconditional branches differ. This dual-KV cache mechanism substantially degrades inference efficiency in modern LLM accelerators and weakens the long-context capabilities of frameworks such as \texttt{vLLM}
% \footnote{https://github.com/vllm-project/vllm}
. Furthermore, CFG originated in the field of diffusion models, where a specific ``null condition'' is typically trained on global data to optimize the unconditional branch. In the context of LLMs, however, CFG lacks this explicit training process, rendering the control capability of its unconditional branch inherently unstable.

To address these challenges, as shown in Fig.~\ref{fig:method}b, we propose to reuse the KV cache of the conditional branch and append a short negative prompt. This KV reuse mechanism eliminates the need for a separate unconditional KV cache, thereby significantly reducing the memory footprint. To tackle the instability of the unconditional branch space, we draw inspiration from negative prompting in diffusion model and construct a negative unconditional branch to steer the model away from erroneous tokens. Specifically, we inject short negative cues (e.g., ``\texttt{OUTPUT ERROR}'') into the reused KV cache, forcing the unconditional branch to generate an undesirable probability distribution. This distribution is then used to perturb high-entropy nodes in the conditional branch, thereby refining the generation results. While we use ``\texttt{OUTPUT ERROR}'' as a general-purpose cue, this prompt can be further tailored to specific tasks for even greater precision. Nevertheless, our experiments show that even this simple two-word cue yields robust performance gains across various datasets.

% \begin{figure*}[!t]
%   \centering
%   \begin{subfigure}{0.24\linewidth}
%     \centering\includegraphics[width=\linewidth]{latex/imgs/nt_word_cloud_c.jpg}
%     \caption{Base (Before-MTI).}
%   \end{subfigure}\hfill
%   \begin{subfigure}{0.24\linewidth}
%     \centering\includegraphics[width=\linewidth]{latex/imgs/nt_word_cloud_cfg.jpg}
%     \caption{Base (After-MTI).}
%   \end{subfigure}\hfill
%   \begin{subfigure}{0.24\linewidth}
%     \centering\includegraphics[width=\linewidth]{latex/imgs/t_word_cloud_c.jpg}
%     \caption{Reasoning (Before-MTI).}
%   \end{subfigure}\hfill
%   \begin{subfigure}{0.24\linewidth}
%     \centering\includegraphics[width=\linewidth]{latex/imgs/t_word_cloud_cfg.jpg}
%     \caption{Reasoning (After-MTI).}
%   \end{subfigure}

%   \caption{Word clouds on GPQA-Diamond comparing Qwen3-8B. We consider the subset of questions that are incorrect under direct inference but become correct after applying our method. Words are extracted from the generated answers, and we visualize only words which the CFG is invoked; word size is proportional to frequency.}
%   \label{fig:word clouds}
% \end{figure*}

\section{Experiments}

\subsection{Experimental setup}
\noindent\textbf{Datasets. }
To assess robustness across diverse domains, we evaluate on three categories of benchmarks: general tasks (MMLU-Pro~\citep{mmlupro}); coding tasks (HumanEval~\citep{humaneval}, HumanEvalPlus, LiveCodeBench~\cite{livecodebench}); and STEM tasks (GPQA-Diamond~\citep{gpqa}, MATH500~\citep{math500}, AIME2024). We use \texttt{OpenCompass}\footnote{https://github.com/open-compass/opencompass}~\citep{opencompass} for consistent and fair evaluation; the corresponding test configuration files and all testing details are provided in Appendix~\ref{sec:config_files}.

\noindent\textbf{Model selection. }
To evaluate the effectiveness of our method, we experiment with the Qwen3~\citep{qwen3} model family (Qwen3-8B, Qwen3-14B, and Qwen3-32B), DeepSeek-R1-7B~\citep{deepseek} and Ling-mini-2.0~\citep{Ling}. 
% Experimental results for the base (non-thinking) versions of the Qwen3 model family are presented in Sec.~\ref{sec:more_results}.

\noindent\textbf{Baselines. }
We compare against the following baselines: (1) Direct Inference (DI): generate outputs directly from the LLM; (2) Vanilla CFG with negative prompt (VC): classifier-free guidance with a specified negative prompt, set to the same prompt injected in our method. Following vanilla CFG~\citep{cfg_llm}, we set $\omega$ to 1.5 and use it as the default hyperparameter. Additionally, a detailed comparison between MTI and current SOTA test-time scaling methods is provided in Tab.~\ref{tab:comparison_with_SOTA_methods}.

\noindent\textbf{Implementation details. }
All experiments set context length of 32768 to ensure that the models' full capabilities are preserved in all datasets. 
To evaluate the effectiveness of MTI 
% as a novel decoding strategy 
and to further examine its ability to correct the model’s probability distribution—where changes to the top-1 prediction are the most directly observable—while ensuring reproducibility, we uniformly adopt greedy search as the base decoding strategy in Tab.~\ref{tab:main_results}.
Additionally, to validate the robustness of our method under stochastic decoding, we evaluate it on AIME2024 using random sampling with temperature 0.6, top-p 0.95, and top-k 20. We run eight independent tests and report the average performance in Tab.~\ref{tab:aime2024}.

\subsection{Main Results}

\begin{table*}[!t]
\centering
\renewcommand{\arraystretch}{1.15}
% \resizebox{0.8\linewidth}{!}{
\resizebox{0.85\linewidth}{!}{
\begin{tabular}{ccccccccccc}
\toprule
% \multirow{2}{*}{\textbf{Model}} & \multirow{2}{*}{\textbf{Method}} & \multirow{2}{*}{\makecell{$\mathbf{\tau}$}} & 
% \multicolumn{2}{c}{\textbf{General}} &
% \multicolumn{2}{c}{\textbf{Math \& Sci}} &
% \multicolumn{3}{c}{\textbf{Coding}} &
% \multirow{2}{*}{\textbf{Avg}} &
% \multirow{2}{*}{\textbf{\makecell{CFG \\ Usage}}} \\
% \cmidrule(lr){4-5} \cmidrule(lr){6-7} \cmidrule(lr){8-10}
%  & & & \textbf{WinoG} & \textbf{MMLU-P} & \textbf{GPQA-D} & \textbf{MATH500} & \textbf{HEval} & \textbf{HEval+} & \textbf{LiveCodeBench} & & \\
\textbf{Model} &
\textbf{Method} &
\makecell{$\mathbf{\tau}$} &
\textbf{MMLU-P} &
\textbf{GPQA-D} &
\textbf{MATH500} &
\textbf{HEval} &
\textbf{HEval+} &
\textbf{LiveCodeBench} &
\textbf{Avg.} &
\textbf{\makecell{CFG \\ Usage}} \\
 
\midrule
\multirow{7}{*}{\rotatebox{90}{\makebox[0pt][c]{\textbf{DeepSeek-R1-7B}}}}
& DI   & -   & 44.45 & 29.29 & 84.60 & 79.27 & 71.34 & 42.75/80.17 & 61.70 & - \\
& VC   & -   & 47.76 & 32.32 & 92.40 & 81.10 & 73.78 & 47.75/{81.21} & 65.19 & 100\% \\
& \multirow{5}{*}{Ours}  
        &   \cc 0.1 & \cc 53.89 & \cc 45.96 & \cc 92.80 & \cc \textbf{89.02} & \cc {78.05} & \cc 51.75/79.96 & \cc 70.20 & \cc 31.4\% \\
        & & \cc 0.5 & \cc \textbf{54.29} & \cc \textbf{51.52} & \cc \textbf{93.60} & \cc {85.98} & \cc 77.44 & \cc \textbf{54.25}/79.75 & \cc \textbf{70.98} & \cc 21.8\% \\
        & & \cc 1.0 & \cc 53.47 & \cc {48.48} & \cc 91.80 & \cc 81.71 & \cc \textbf{81.71} & \cc {52.00}/79.75 & \cc 69.85 & \cc 12.7\% \\
        & & \cc 1.5 & \cc 52.31 & \cc 40.91 & \cc {93.20} & \cc 81.71 & \cc 73.17 & \cc 50.50/80.79 & \cc 67.51 & \cc 9.3\% \\
        & & \cc 2.0 & \cc 50.41 & \cc 38.89 & \cc 90.60 & \cc 80.49 & \cc 75.61 & \cc 48.00/\textbf{81.84} & \cc 66.55 & \cc 3.6\% \\

\midrule

\multirow{7}{*}{\rotatebox{90}{\makebox[0pt][c]{\textbf{Ling-mini-2.0}}}}
& DI   & -   & 63.93 & 52.02 & 94.00 & 87.20 & 75.00 & 57.00/38.62 & 66.82 & - \\
& VC   & -   & 63.48 & 53.54 & \textbf{95.80} & 85.37 & 78.66 & 56.50/40.29 & 67.66 & 100\% \\
& \multirow{5}{*}{Ours}
        & \cc 0.1 & \cc 63.77 & \cc \textbf{56.57} & \cc 94.80 & \cc \textbf{87.80} & \cc 78.66 & \cc 56.50/{42.80} & \cc \textbf{68.70} & \cc 28.2\% \\
        & & \cc 0.5 & \cc \textbf{64.54} & \cc 52.02 & \cc 94.80 & \cc 86.59 & \cc \textbf{79.27} & \cc 57.25/\textbf{45.09} & \cc {68.51} & \cc 14.8\% \\
        & & \cc 1.0 & \cc 63.86 & \cc 52.02 & \cc 95.00 & \cc 86.59 & \cc 76.22 & \cc 55.00/41.75 & \cc 67.21 & \cc 4.5\% \\
        & & \cc 1.5 & \cc 63.18 & \cc {54.04} & \cc 94.60 & \cc 87.20 & \cc 76.22 & \cc {57.50}/39.46 & \cc 67.46 & \cc 0.9\% \\
        & & \cc 2.0 & \cc 63.81 & \cc 51.52 & \cc 95.20 & \cc 87.20 & \cc 75.00 & \cc \textbf{58.00}/39.67 & \cc 67.20 & \cc 0.2\% \\
\midrule

\multirow{7}{*}{\rotatebox{90}{\makebox[0pt][c]{\textbf{Qwen3-8B}}}}
& DI   & -   & 70.52 & 57.07 & 96.80 & 87.20 & 67.07 & 79.50/97.70 & 79.41 & - \\
& VC   & -   & 70.19 & 57.58 & 92.60 & 90.85 & 65.85 & 84.75/93.74 & 79.37 & 100\% \\
& \multirow{5}{*}{Ours}
            & \cc 0.1 & \cc 70.69 & \cc 54.55 & \cc 96.80 & \cc {92.07} & \cc \textbf{74.39} & \cc \textbf{88.00}/96.45 & \cc 81.85 & \cc 50.5\% \\
            & & \cc 0.5 & \cc 71.91 & \cc 60.61 & \cc \textbf{97.00} & \cc \textbf{95.12} & \cc {69.51} & \cc 87.25/\textbf{98.12} & \cc \textbf{82.79} & \cc 30.5\% \\
            & & \cc 1.0 & \cc {72.28} & \cc {60.10} & \cc \textbf{97.00} & \cc 90.85 & \cc 67.07 & \cc \textbf{88.00}/{97.91} & \cc {81.89} & \cc 9.8\% \\
            & & \cc 1.5 & \cc \textbf{72.38} & \cc \textbf{61.11} & \cc 95.80 & \cc 89.02 & \cc 65.24 & \cc 84.75/97.49 & \cc 80.83 & \cc 3.3\% \\
            & & \cc 2.0 & \cc 70.96 & \cc 56.06 & \cc 96.80 & \cc 88.41 & \cc 65.85 & \cc 81.50/97.70 & \cc 79.61 & \cc 0.4\% \\
\midrule
\multirow{7}{*}{\rotatebox{90}{\makebox[0pt][c]{\textbf{Qwen3-14B}}}}
& DI   & -   & 75.61 & 57.58 & 96.40 & 92.07 & 70.12 & 86.00/{98.54} & 82.33 & - \\
& VC   & -   & 74.12 & 60.10 & 94.80 & 90.85 & 69.51 & 89.00/\textbf{98.96} & 82.48 & 100\% \\
& \multirow{5}{*}{Ours}  
            & \cc 0.1 & \cc 74.86 & \cc \textbf{64.14} & \cc 97.00 & \cc {95.73} & \cc 72.56 & \cc 90.25/97.91 & \cc 84.64 & \cc 48.6\% \\
            & & \cc 0.5 & \cc 74.63 & \cc \textbf{64.14} & \cc {97.20} & \cc \textbf{96.95} & \cc {75.00} & \cc \textbf{91.50}/97.91 & \cc \textbf{85.33} & \cc 32.8\% \\
            & & \cc 1.0 & \cc \textbf{76.12} & \cc 62.12 & \cc 97.00 & \cc 95.12 & \cc \textbf{75.61} & \cc {90.50}/97.49 & \cc {84.85} & \cc 10.3\% \\
            & & \cc 1.5 & \cc {75.91} & \cc 62.12 & \cc \textbf{97.60} & \cc {95.73} & \cc 69.51 & \cc 88.50/97.70 & \cc 83.87 & \cc 3.5\% \\
            & & \cc 2.0 & \cc 75.64 & \cc 61.62 & \cc {97.20} & \cc 93.90 & \cc 70.73 & \cc 88.00/98.33 & \cc 83.63 & \cc 0.3\% \\
\midrule
\multirow{7}{*}{\rotatebox{90}{\makebox[0pt][c]{\textbf{Qwen3-32B}}}}
& DI   & -   & 76.76 & 61.62 & 97.00 & 97.56 & 97.56 & 91.00/\textbf{99.16} & 88.67 & - \\
& VC   & -   & 74.96 & 56.57 & 94.00 & 95.12 & 95.73 & 87.25/98.54 & 86.02 & 100\% \\
& \multirow{5}{*}{Ours}  
            & \cc 0.1 & \cc 76.56 & \cc 64.65 & \cc \textbf{98.40} & \cc 96.34 & \cc 97.56 & \cc \textbf{92.00}/98.75 & \cc 89.18 & \cc 49.1\% \\
            & & \cc 0.5 & \cc 77.15 & \cc \textbf{65.66} & \cc \textbf{98.40} & \cc \textbf{98.17} & \cc \textbf{98.17} & \cc 90.75/98.54 & \cc {89.55} & \cc 31.2\% \\
            & & \cc 1.0 & \cc \textbf{78.47} & \cc \textbf{65.66} & \cc 97.40 & \cc \textbf{98.17} & \cc 97.56 & \cc {91.25}/\textbf{99.16} & \cc \textbf{89.67} & \cc 13.1\% \\
            & & 1.5 & {77.80} & 60.61 & 96.20 & 97.56 & 97.56 & 89.75/98.75 & 88.32 & 3.4\% \\
            & & 2.0 & 76.86 & 61.62 & 96.00 & 96.34 & \textbf{98.17} & 90.00/98.75 & 88.25 & 0.7\% \\

\bottomrule
\end{tabular}
}
\caption{Performance comparison on general, coding and STEM benchmarks. We compare Direct Inference (DI), Vanilla CFG (VC), and Ours (MTI) under varying entropy thresholds ($\tau$). Avg. (\%) and CFG Usage denote the average performance and CFG usage ratio, respectively. Green cells indicate cases where Ours outperforms DI, and Ours consistently surpasses both DI and VC.}
\label{tab:main_results}
\end{table*}

% (44.45 + 29.29 + 84.60 + 79.27 + 71.34 + 42.75 + 80.17)/7=
% (47.76 + 32.32 + 92.40 + 81.10 + 73.78 + 47.75 + 81.21)/7=
% (53.89 + 45.96 + 92.80 + 89.02 + 78.05 + 51.75 + 79.96)/7=
% (54.29 + 50.00 + 94.00 + 85.98 + 76.83 + 54.25 + 79.75)/7=
% (53.47 + 48.48 + 91.80 + 81.71 + 81.71 + 52.00 + 79.75)/7=
% (52.31 + 40.91 + 93.20 + 81.71 + 73.17 + 50.50 + 80.79)/7=
% (50.41 + 38.89 + 90.60 + 80.49 + 75.61 + 48.00 + 81.84)/7=

Experimental results in Tab.~\ref{tab:main_results} and Tab.~\ref{tab:aime2024} show that MTI consistently outperforms Direct Inference (DI) and Vanilla CFG (VC). Notably, MTI exhibits a broad ``green zone'' where it surpasses DI across nearly all threshold configurations, proving its robustness without meticulous tuning. 

As shown in Tab.~\ref{tab:main_results}, on DeepSeek-R1-7B, MTI reaches an average score of 70.73\% (+9.28\% over DI) with 21.8\% CFG usage, and on Qwen3-14B, MTI reaches an average score of 85.33\% (+3.00\% over DI) with 32.8\% CFG usage. Crucially, MTI rectifies erroneous distributions where traditional strategies fail. For instance, unlike rank-preserving methods such as temperature sampling, MTI reshapes the distribution by re-ranking logits to elevate suppressed candidates. 
For example, on GPQA-Diamond with DeepSeek-R1-7B, DI with greedy decoding leads to highly repetitive generations, resulting in a poor score of 29.29\%. By contrast, MTI corrects the erroneous top-1 logits, improving performance to 51.52\% (+22.23\%).

On AIME2024 (Tab.~\ref{tab:aime2024}), MTI demonstrates its effectiveness under random sampling, boosting DeepSeek-R1-7B from 54.17\% to 62.92\% (+8.75\%) and Ling-mini-2.0 from 60.00\% to 71.25\% (+11.25\%). In contrast, VC often underperforms DI in AIME2024 (e.g., 73.34\% vs. 73.75\% on Qwen3-8B). As analyzed in Sec.~\ref{sec:different_entropy_scopes}, applying CFG in low-entropy scopes ($\le 1.5$) actually degrades performance to 71.67\%, whereas targeting high-entropy states ($> 1.5$) yields 78.34\%. This suggests that in long-form reasoning, excessive intervention during high-confidence steps disturbs the model’s logical flow, leading to cumulative errors. By bypassing these states, MTI mitigates error propagation and achieves competitive performance against SOTA methods in Tab.~\ref{tab:comparison_with_SOTA_methods}. 
% As a plug-and-play framework, MTI is orthogonal to and can be integrated into existing test-time scaling frameworks to further enhance reasoning fidelity. 
These results validate MTI as a scalable and reliable approach for enhancing LLM reasoning without the need for exhaustive hyperparameter search.

\begin{figure*}[t]
    \centering
    \begin{subfigure}[b]{0.32\textwidth}
        \centering
        \includegraphics[width=\linewidth]{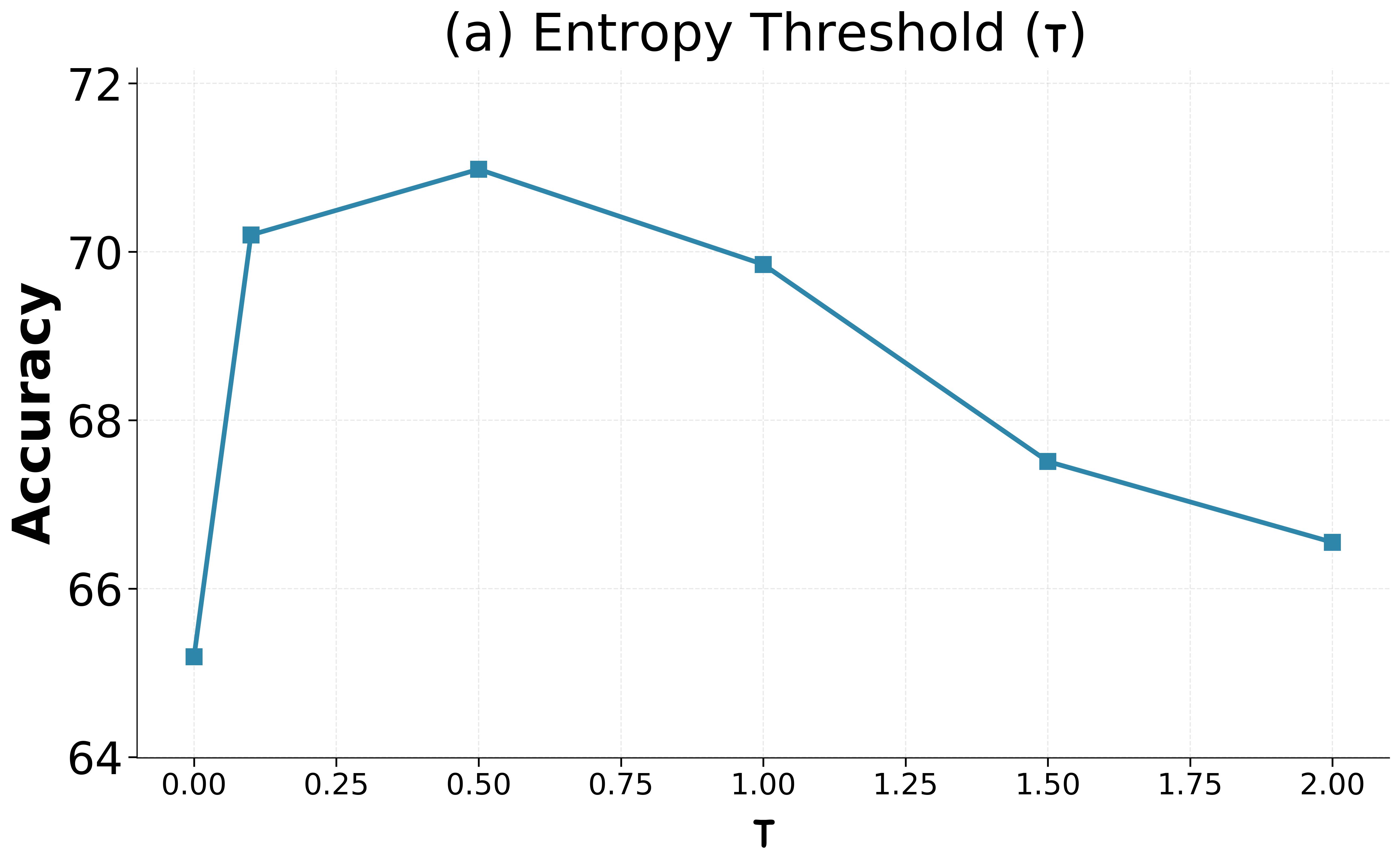}
        % \subcaption{}
        % \label{fig:ablations_et}
    \end{subfigure}
    \hfill
    \begin{subfigure}[b]{0.32\textwidth}
        \centering
        \includegraphics[width=\linewidth]{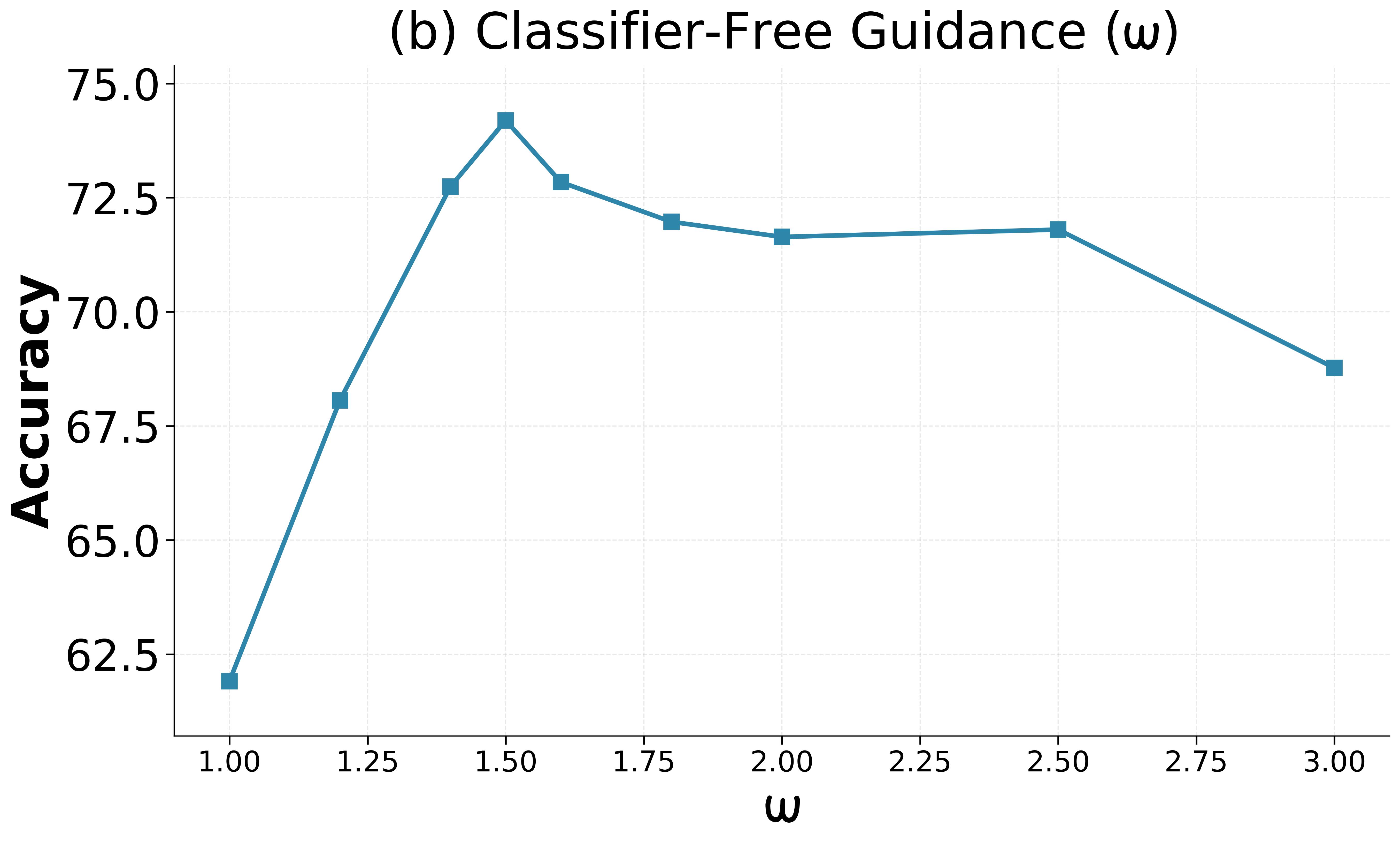}
        % \subcaption{}
        % \label{fig:ablations_w}
    \end{subfigure}
    \hfill
    \begin{subfigure}[b]{0.32\textwidth}
        \centering
        \includegraphics[width=\linewidth]{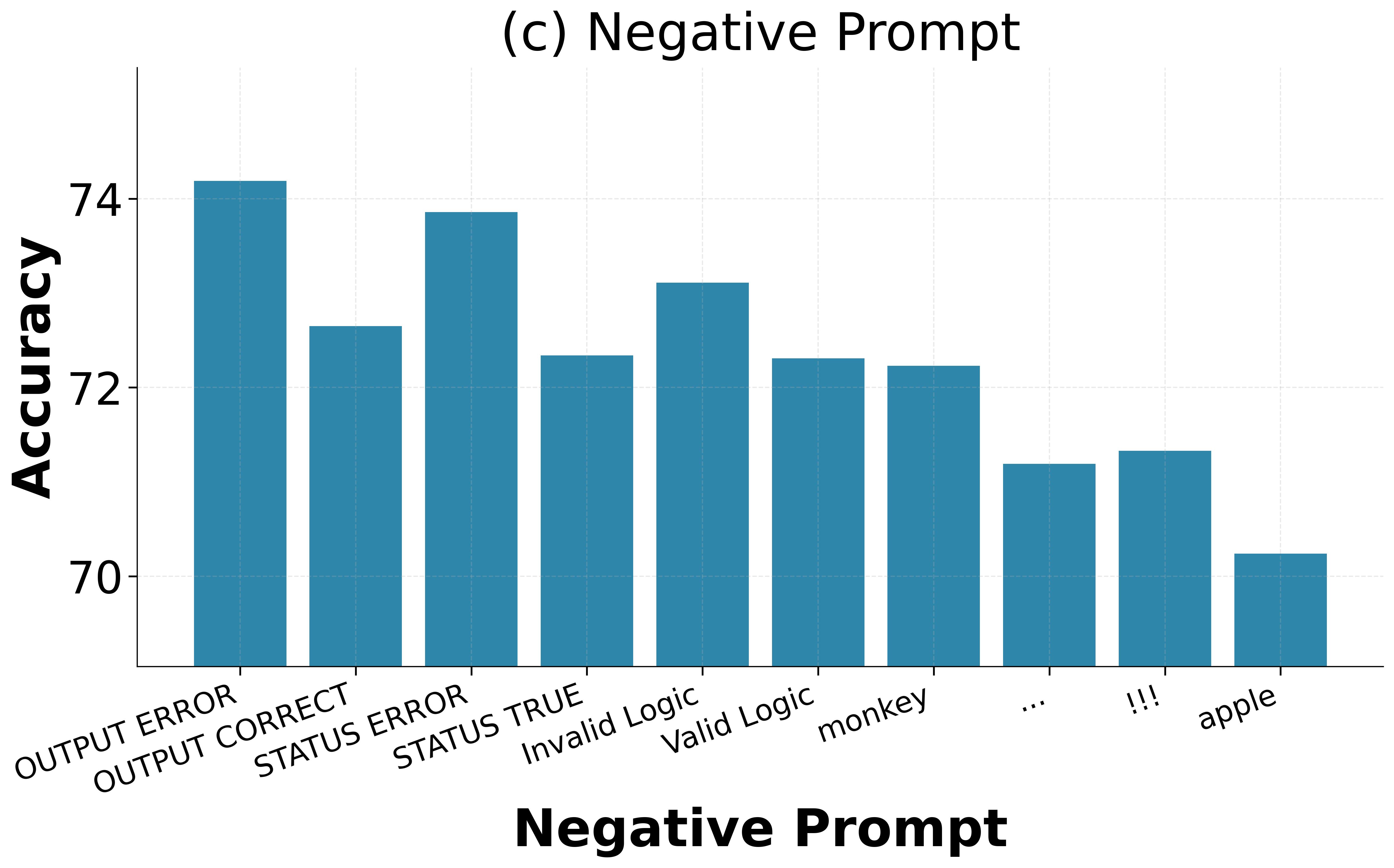}
        % \subcaption{}
        % \label{fig:ablations_np}
    \end{subfigure}
    \caption{Ablation study results on the selection of (a) entropy threshold, (b) $\omega$ in classifier-free guidance and (c) negative prompt in MTI.}
    \label{fig:ablations}
\end{figure*}

% \begin{table}[!t]
% \centering
% \renewcommand{\arraystretch}{1.15}
% \setlength{\tabcolsep}{6pt}
% \resizebox{1.0\linewidth}{!}{

% \begin{tabular}{cc*{6}{c}}
% \toprule
% \multirow{1}{*}{\textbf{Method}} & \multirow{1}{*}{\makecell{$\mathbf{\tau}$}} &
% \multicolumn{1}{c}{\textbf{Qwen3-8B}} & 
% \multicolumn{1}{c}{\textbf{Qwen3-14B}} & 
% \multicolumn{1}{c}{\textbf{Qwen3-32B}} \\ 
% \midrule 
% DI & -- & 73.75 / -- & 78.33 / -- & 79.58 / -- \\
% VC & -- & 73.34 / 100\% & 75.42 / 100\% & 72.92 / 100\% \\
% \midrule
% \multirow{5}{*}{Ours} 
% & 0.1 & 76.60 / 37.8\% & 79.58 / 40.0\% & \underline{82.92} / 41.7\% \\
% & 0.5 & \underline{77.08} / 22.5\% & \underline{80.83} / 24.4\% & \textbf{84.58} / 25.7\% \\
% & 1.0 & 75.42 / 8.0\% & \textbf{81.67} / 8.2\% & \underline{82.92} / 9.0\%\\
% & 1.5 & \textbf{78.34} / 1.9\% & 78.33 / 2.0\% & 82.50 / 2.2\%\\
% & 2.0 & 75.00 / 0.3\% & 80.00 / 0.3\% & 77.92 / 0.3\%\\
% \bottomrule
% \end{tabular}

\begin{table}[t]
\centering
\renewcommand{\arraystretch}{1.15}
\setlength{\tabcolsep}{6pt}
\resizebox{1.0\linewidth}{!}{

\begin{tabular}{cc*{8}{c}}
\toprule
\multirow{1}{*}{\textbf{Method}} & \multirow{1}{*}{\makecell{$\mathbf{\tau}$}} &
\multicolumn{1}{c}{\textbf{DS-R1-7B}} &
\multicolumn{1}{c}{\textbf{Ling-mini-2.0}} &
\multicolumn{1}{c}{\textbf{Qwen3-8B}} & 
\multicolumn{1}{c}{\textbf{Qwen3-14B}} & 
\multicolumn{1}{c}{\textbf{Qwen3-32B}} & \\
\midrule 
DI & -- & 54.17 / -- & 60.00 / -- & 73.75 / -- & 78.33 / -- & 79.58 / -- \\
VC & -- & 48.33 / 100\% & 65.00 / 100\% & 73.34 / 100\% & 75.42 / 100\% & 72.92 / 100\% \\
\midrule
\multirow{5}{*}{Ours} 
& 0.1 & \cellcolor{mycolor} 57.08 / 51.6\% & \cellcolor{mycolor} 65.00 / 15.8\% & \cellcolor{mycolor} 76.60 / 37.8\% & \cellcolor{mycolor} 79.58 / 40.0\% & \cellcolor{mycolor} \underline{82.92} / 41.7\% \\
& 0.5 & 51.67 / 36.9\% & \cellcolor{mycolor} \underline{68.34} / 7.6\% & \cellcolor{mycolor} \underline{77.08} / 22.5\% & \cellcolor{mycolor} \underline{80.83} / 24.4\% & \cellcolor{mycolor} \textbf{84.58} / 25.7\% \\
& 1.0 & \cellcolor{mycolor} \textbf{62.92} / 21.1\% & \cellcolor{mycolor} \textbf{71.25} / 1.4\% & \cellcolor{mycolor} 75.42 / 8.0\% & \cellcolor{mycolor} \textbf{81.67} / 8.2\% & \cellcolor{mycolor} \underline{82.92} / 9.0\% \\
& 1.5 & \cellcolor{mycolor} \underline{60.00} / 13.2\% & \cellcolor{mycolor} 63.33 / 0.2\% & \cellcolor{mycolor} \textbf{78.34} / 1.9\% & 78.33 / 2.0\% & \cellcolor{mycolor} 82.50 / 2.2\% \\
& 2.0 & \cellcolor{mycolor} 58.75 / 6.9\% & \cellcolor{mycolor} 65.00 / 0.01\% & \cellcolor{mycolor} 75.00 / 0.3\% & \cellcolor{mycolor} 80.00 / 0.3\% & 77.92 / 0.3\% \\
\bottomrule
\end{tabular}
}
\caption{
AIME2024 accuracy (\%) and CFG usage for different models and methods. Results are averaged over 8 random sampling runs. Green cells indicate cases where Ours (MTI) outperforms DI, and Ours consistently surpasses both DI and VC.}
\label{tab:aime2024}
\end{table}

\subsection{Ablation Study}
\label{sec:Ablation Study}

\noindent\textbf{Entropy threshold $\tau$. } Fig.~\ref{fig:ablations}a shows the ablation study on $\tau$, with the curve representing the average accuracy from DeepSeek-R1-7B results in Tab.~\ref{tab:main_results}. As $\tau$ increases, performance follows a unimodal pattern, rising initially before declining. A low $\tau$ over-modifies low-entropy tokens, destabilizing the reasoning chain, while a high $\tau$ leaves high-entropy tokens uncorrected, degrading performance. Additionally, the ``green zone'' in Tab.~\ref{tab:main_results} reveals that MTI consistently outperforms DI across most threshold configurations, demonstrating robustness without extensive tuning. As shown in Tab.~\ref{tab:main_results} and Sec.~\ref{tab:comparison_with_SOTA_methods}, at an entropy threshold of 0.5, MTI outperforms all comparison methods. For selecting the optimal $\tau$, cross-validation on a held-out dataset can be used.

\noindent\textbf{CFG hyperparameters $\omega$. }
Fig.~\ref{fig:ablations}b investigates the impact of $\omega$ on the average accuracy of DeepSeek-R1-7B across the GPQA-Diamond, Math500, and HumanEvalPlus datasets, following the experimental settings in Tab.~\ref{tab:main_results}, where the performance exhibits an initial increase followed by a subsequent decline. The model attains its peak accuracy when $\omega=1.5$. Notably, the performance is fairly stable around this value, indicating that $\omega$ is not sensitive and typically does not require extensive tuning.

\noindent\textbf{Negative prompt.}
Fig.~\ref{fig:ablations}c shows the impact of negative prompts on the average accuracy of DeepSeek-R1-7B across the GPQA-Diamond, Math500, and HumanEvalPlus datasets. Experimental results reveal two key observations. First, negative prompts (e.g., \texttt{OUTPUT ERROR}, \texttt{Invalid Logic}) consistently outperform positive prompts (e.g., \texttt{OUTPUT CORRECT}, \texttt{Valid Logic}). Second, negative prompts that are semantically grounded perform significantly better than meaningless or weakly informative tokens (e.g., \texttt{apple}, \texttt{!!!}). These findings confirm that, within the MTI framework, negative prompts effectively guide the model away from undesirable conditional distributions. By explicitly encoding errors or logical failures, these prompts suppress the generation probability of incorrect tokens more reliably. In contrast, meaningless tokens introduce only unspecific perturbations, which makes them less effective. While task-specific negative prompts can yield additional improvements (Sec.~\ref{sec:task_specific_np}), \texttt{OUTPUT ERROR} demonstrates robustness, making it a general-purpose heuristic across tasks.

\subsection{Analysis}

\begin{figure}[h]
  \centering
  \includegraphics[width=\linewidth]{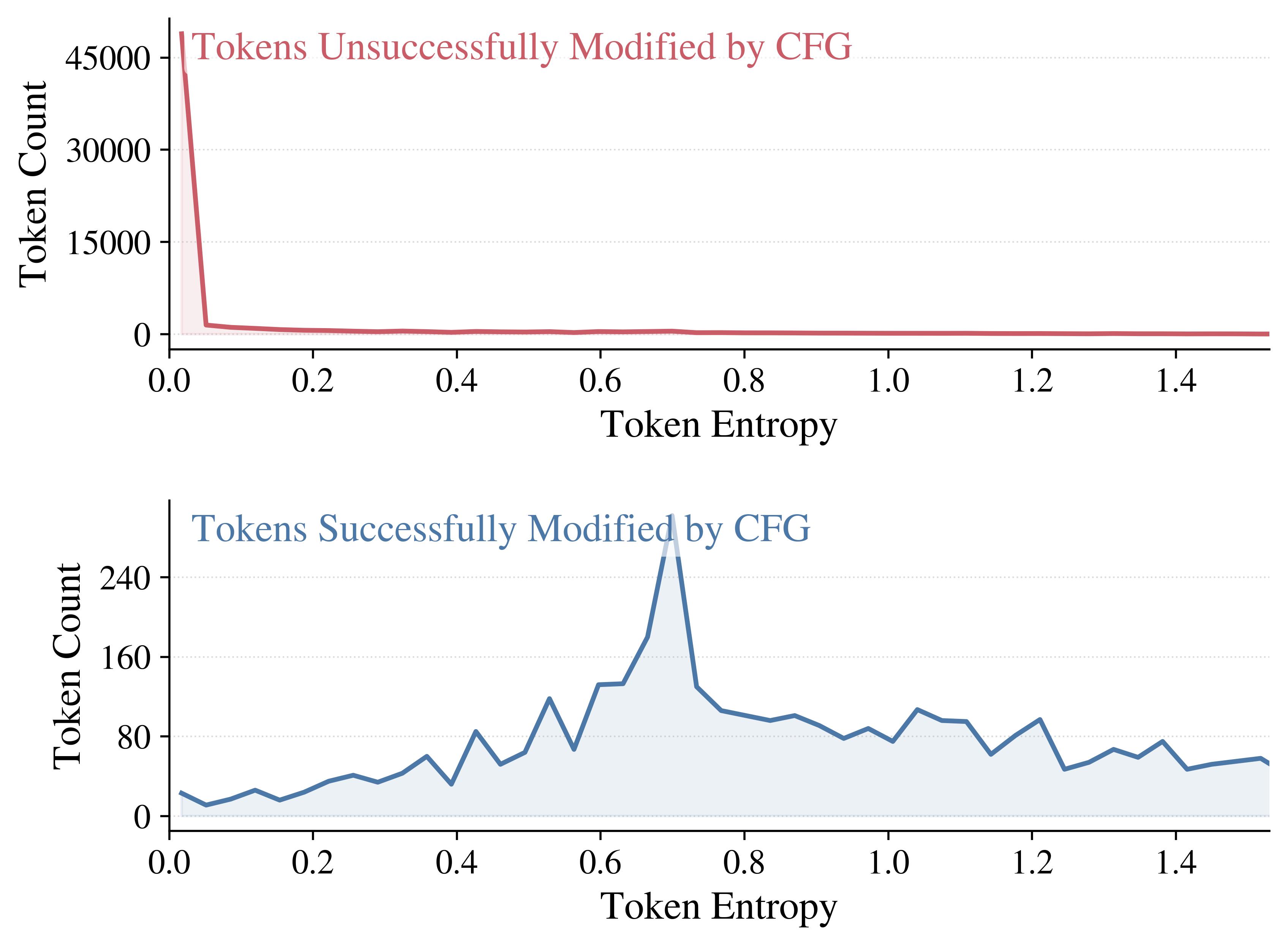}
  % \caption {Results on HumanEvalPlus using Qwen3-8B with vanilla CFG. Top: tokens where CFG failed to change the prediction; bottom: tokens successfully changed. x-axis: token entropy; y-axis: count of tokens per entropy bin.}
  \caption{Entropy distributions of tokens categorized by vanilla CFG modification success. x-axis: token entropy; y-axis: token count per bin.}
  \label{fig:motivation_figure}
\end{figure}

\noindent\textbf{Not all tokens are created equal in CFG. }To investigate the intrinsic mechanism and provide a principled justification for our method, we conduct a greedy search experiment using 
% Qwen3-8B-Base 
Qwen3-8B 
on the HumanEvalPlus dataset, as illustrated in Fig.~\ref{fig:motivation_figure}. We specifically analyze the tokens generated during the inference phase, focusing on the relationship between their logits entropy and the effectiveness of CFG in altering the top-1 prediction. The empirical results reveal a clear divergence: as shown in the top panel of Fig.~\ref{fig:motivation_figure}, the vast majority of tokens remain unchanged after applying CFG, with these instances predominantly concentrated in the low-entropy region. Conversely, the bottom panel demonstrates that tokens whose predictions are successfully steered by CFG are primarily those with high entropy. We attribute this phenomenon to the confidence levels inherent in the logits distribution. Low-entropy logits typically contain a single dominant mode, indicating that the model is already highly confident in its prediction; this makes the token naturally resistant to CFG-based modification. These observations resonate with our core motivation: since low-entropy tokens are often ``immune'' to CFG, it is both reasonable and efficient to apply CFG selectively to high-entropy tokens, thereby avoiding redundant computations on predictions that are unlikely to change.

\begin{figure}[h]
  \centering
  \begin{subfigure}{0.48\linewidth}
    \centering\includegraphics[width=\linewidth]{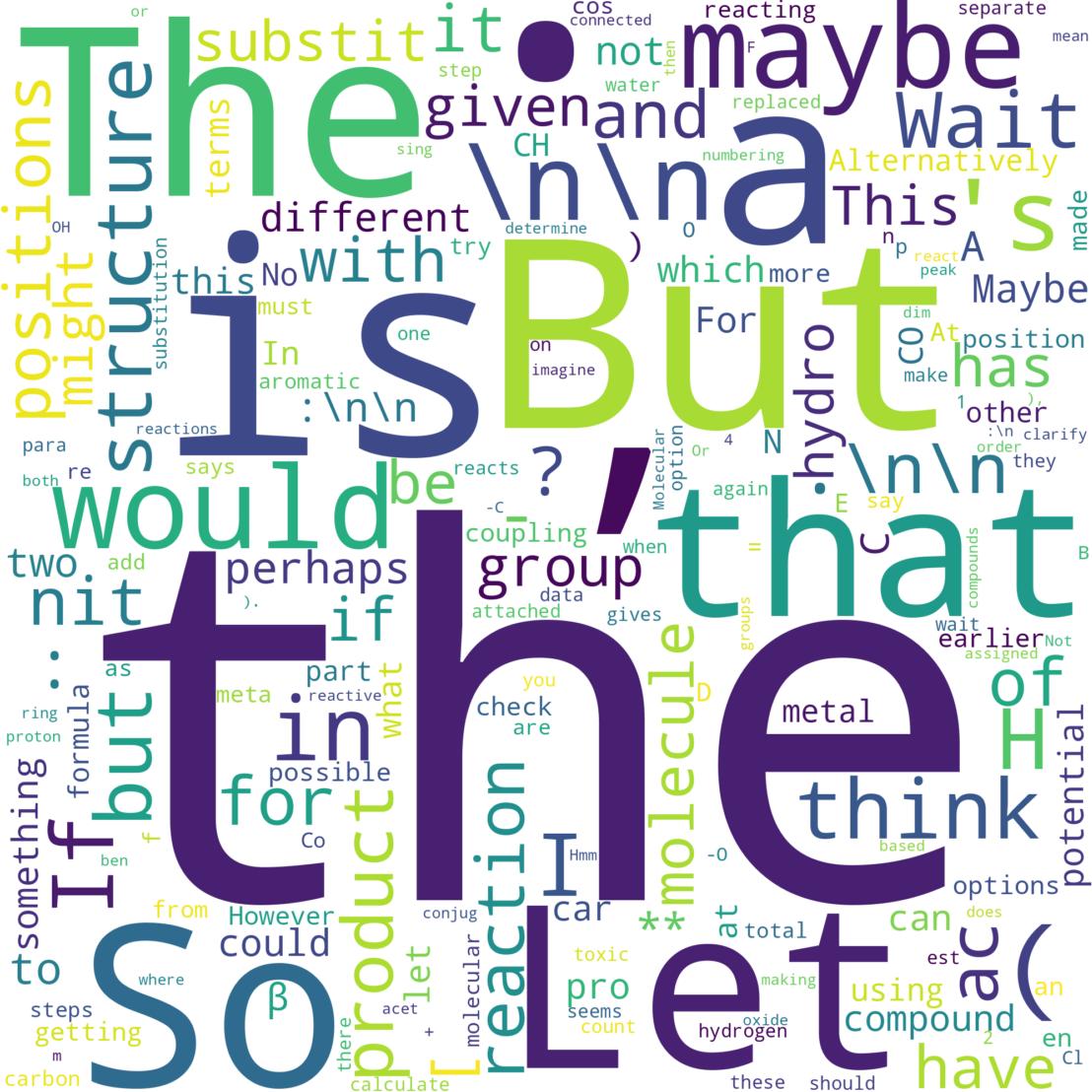}
    \caption{Before-MTI.}
    % \label{fig:word_cloud_before}
  \end{subfigure}\hfill
  \begin{subfigure}{0.48\linewidth}
    \centering\includegraphics[width=\linewidth]{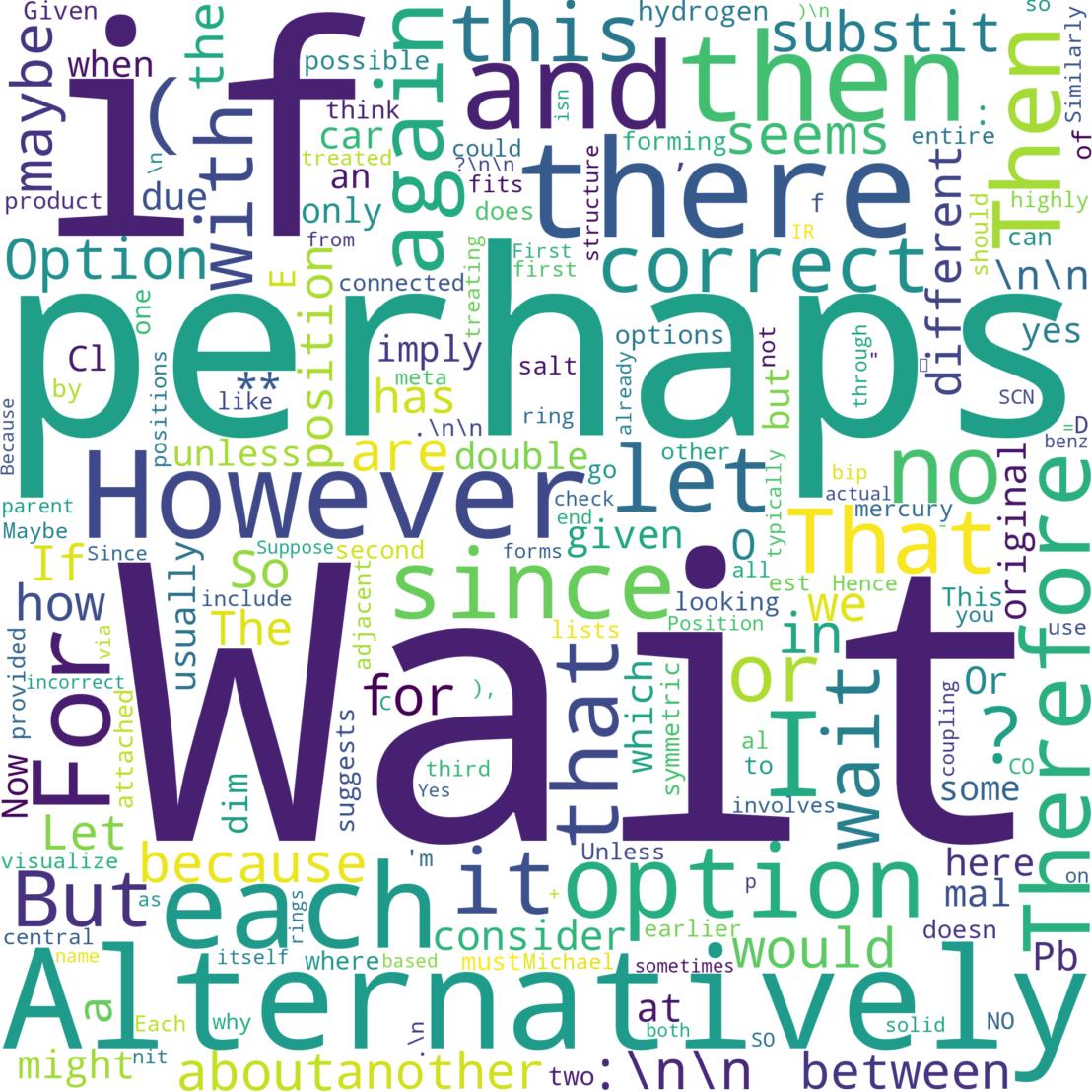}
    \caption{After-MTI.}
    % \label{fig:word_cloud_after}
  \end{subfigure}
  \caption{Token word clouds before vs. after MTI.}
  \label{fig:word clouds}
\end{figure}

\noindent\textbf{Word-cloud analysis.} As shown in Fig.~\ref{fig:word clouds}, we conduct a word cloud visualization on Qwen3-8B using the GPQA-Diamond dataset, specifically focusing on questions where the model initially fail but reach the correct answer after applying MTI. By comparing the high entropy token distributions before and after intervention, we observe that MTI significantly enhances lexical diversity: generic tokens (e.g., the, a) are frequently replaced by semantically dense reasoning terms and connectives. Notably, there is a distinct shift away from simple sequential connectives (e.g., so) toward more diverse and critical connectives, such as adversative conjunctions and hedging terms (e.g., however, if, perhaps, alternatively, wait). This transition suggests that MTI helps the model abandon erroneous prior reasoning chains and initiate new lines of thought, effectively steering the trajectory toward correct inferences. Overall, applying MTI yields more varied and balanced vocabularies, which expand the hypothesis space for reasoning trajectories and ultimately result in better outcomes.

\begin{table}[h]
\centering
\renewcommand{\arraystretch}{1.15}
\setlength{\tabcolsep}{6pt}
% \resizebox{0.85\linewidth}{!}{
\resizebox{1.0\linewidth}{!}{
\begin{tabular}{cccccc}
    \hline
    \textbf{Model} & \textbf{Method} & $\mathbf{\tau}$ & \textbf{Accuracy (\%)} & \textbf{Latency} \\
    \hline
    \multirow{6}{*}{\rotatebox{90}{\makebox[0pt][c]{\textbf{Qwen3-14B}}}}
    & \textbf{DI}    & -- & 78.262 & 4547s \\
    & \textbf{VC}    & -- & 76.296 & 12411s \\
    & \multirow{4}{*}{\textbf{Ours}}
    &   0.1 & 79.082 & 6081s \\
    & & 0.5 & 80.660 & 5289s \\
    & & 1.0 & \textbf{80.698} & \textbf{4551s} \\
    & & 1.5 & 80.052 & 3954s \\
    % & & 2.0 & 79.812 & 4227s \\
    \hline
    \multirow{6}{*}{\rotatebox{90}{\makebox[0pt][c]{\textbf{Ling-mini-2.0}}}}
    & \textbf{DI}    & -- & 74.556 & 1288s \\
    & \textbf{VC}    & -- & 75.460 & 3142s \\
    & \multirow{4}{*}{\textbf{Ours}}
    &   0.1 & \textbf{76.400} & \textbf{1495s} \\
    & & 0.5 & 75.338 & 1444s \\
    & & 1.0 & 74.736 & 1370s \\
    & & 1.5 & 75.308 & 1302s \\
    % & & 2.0 & 74.664 & 1058s \\
    \hline
\end{tabular}
}
\caption{Inference Cost. Ours (MTI) consistently outperforms DI and VC with negligible overhead.}
\label{tab:inference_cost}
\end{table}

\noindent\textbf{Inference cost.}
We evaluate the latency of Qwen3-14B and Ling-mini-2.0 on the GPQA-Diamond, Math500, HumanEval, Winogrande, and HumanEvalPlus datasets. Here, latency refers to the total time required to evaluate all datasets on a single GPU following the setting in Tab.~\ref{tab:main_results}. As shown in Tab.~\ref{tab:inference_cost}, we observe that MTI does not necessarily increase the inference time. This is because MTI can guide the reasoning trajectory more effectively and reduce unnecessary or incorrect reasoning chains, resulting in shorter output. This suggests that MTI can spontaneously achieve a token-reduction effect~\citep{kong2025token}.

\subsection{Applying MTI to vision language models}

\begin{table}[h]
\centering
\renewcommand{\arraystretch}{1.15}
\setlength{\tabcolsep}{6pt}
% \resizebox{0.85\linewidth}{!}{
\resizebox{1.0\linewidth}{!}{
    \begin{tabular}{cccc}
        \hline
        \textbf{Method} & \textbf{Where2Place} & \textbf{RefSpatial} & \textbf{VaBenchPoint} \\
        \hline
        \textbf{DI}   & 61.39/-- & 42.45/-- & 38.87/-- \\
        \textbf{VC}    & 61.76/100\% & 39.86/100\% & 35.88/100\% \\
        \textbf{Ours}  & \textbf{64.36}/3.7\% & \textbf{44.24}/3.7\% & \textbf{40.86}/4.5\% \\
        \hline
    \end{tabular}
}
\caption{Evaluation of Ours (MTI) with vision-language model. Results are reported as Accuracy(\%) / CFG Usage. Ours consistently outperforms DI and VC.
% Bold numbers indicate the best performance in accuracy
}
\label{tab:vlm}
\end{table}

To examine whether MTI generalizes across pretraining objectives, we extend it to the vision-language model (VLM) setting. We evaluate MTI on three spatial understanding benchmarks—Where2Place~\citep{Where2Place}, RefSpatial~\citep{RefSpatial}, and VaBenchPoint~\citep{VaBenchPoint}—using Qwen3-8B-VL-Instruct. We strictly follow the official evaluation prompts of Qwen3-8B-VL-Instruct and report results in the same format as Tab.~\ref{tab:vlm}. With an entropy threshold of $\tau=2.0$, MTI is activated for only 3.7\%–4.5\% of generated tokens, yet consistently improves accuracy: 61.39\%$\rightarrow$64.36\%(+2.97\%) on Where2Place, 42.45\%$\rightarrow$44.24\%(+1.79\%) on RefSpatial, and 38.87\%$\rightarrow$40.86\%(+1.99\%) on VaBenchPoint.

% \section{Conclusions And Future Work}
% We find in the ablation studies that different negative prompts have non-negligible impacts on the results. In future work, we will explore how to identify optimal negative prompts and compress longer prompts into a single token to further improve inference efficiency.

\section{Conclusion}

We propose Minimal Test-Time Intervention (MTI), a training-free framework that improves LLM reasoning with minimal overhead. We first introduce selective CFG intervention to use token entropy to identify unstable positions and apply classifier-free guidance only at those tokens, stabilizing the reasoning trajectory without unnecessary computation. We further introduce the lightweight negative-prompt guidance to approximate the unconditional branch without additional memory cost. Experiments across general, coding and STEM benchmarks show consistent improvements. MTI is plug-and-play, compatible with modern acceleration frameworks, providing an effective mechanism for enhancing test-time reasoning reliability.
% and orthogonal to existing test-time scaling methods, 

\section{Limitations}

In our experiments, we find that meaningless tokens, such as ``\texttt{apple}'', as well as positive prompts, improve performance over Direct Inference, though not as effectively as semantically explicit negative prompts like ``\texttt{OUTPUT ERROR}''. The interpretability of this effect remains unexplored, and further investigation into this area is a key direction for future research. Additionally, analyses of failure cases are provided in Sec.~\ref{sec:different_entropy_scopes} and Sec.~\ref{sec:failure_cases_analysis}.

% Bibliography entries for the entire Anthology, followed by custom entries
% \bibliography{anthology,custom}
% Custom bibliography entries only
\bibliography{custom}
\appendix

\section{Appendix}

% \subsection{Answer-level entropy}
% \label{sec:answer_level_entropy}
% $$ A_n= \sum_{t=1}^{T_n}H_t. $$
% $H_t$ denotes token-level entropy in Sec.~\ref{sec:token_level_entropy}. $A_n$ denotes the answer-level entropy of the full reasoning trajectory for the $n$-th question, where $T_n$ represents the length of the full reasoning trajectory for that question.
% $$ A=\frac{1}{N} \sum_{n=1}^N A_n. $$
% $A$ denotes the average answer entropy, where $N$ represents the number of data instances.

\subsection{Testing setup and files}
\label{sec:config_files}
% Add this to your preamble: \usepackage{hyperref}
% You can also add \usepackage[T1]{fontenc} to improve typewriter text.

% \paragraph{OpenCompass configuration files used in our main results.}
For our results, we use the following \texttt{OpenCompass} configuration files:
\begin{itemize}
  \item \href{https://github.com/open-compass/opencompass/blob/main/opencompass/configs/datasets/gpqa/gpqa_gen.py}{\texttt{gpqa\_gen.py}}
  \item \href{https://github.com/open-compass/opencompass/blob/main/opencompass/configs/datasets/math/math_500_gen.py}{\texttt{math\_500\_gen.py}}
  \item \href{https://github.com/open-compass/opencompass/blob/main/opencompass/configs/datasets/aime2024/aime2024_gen.py}{\texttt{aime2024\_gen.py}}
  \item \href{https://github.com/open-compass/opencompass/blob/main/opencompass/configs/datasets/humaneval/humaneval_gen.py}{\texttt{humaneval\_gen.py}}
  \item \href{https://github.com/open-compass/opencompass/blob/main/opencompass/configs/datasets/humaneval_plus/humaneval_plus_gen.py}{\texttt{humaneval\_plus\_gen.py}}
  \item \href{https://github.com/open-compass/opencompass/blob/main/opencompass/configs/datasets/livecodebench/livecodebench_gen.py}{\texttt{livecodebench\_gen.py}}
  % \item \href{https://github.com/open-compass/opencompass/blob/main/opencompass/configs/datasets/hellaswag/hellaswag_gen.py}{\texttt{hellaswag\_gen.py}}
  % \item \href{https://github.com/open-compass/opencompass/blob/main/opencompass/configs/datasets/winogrande/winogrande_gen.py}{\texttt{winogrande\_gen.py}}
  \item \href{https://github.com/open-compass/opencompass/blob/main/opencompass/configs/datasets/mmlu_pro/mmlu_pro_gen.py}{\texttt{mmlu\_pro\_gen.py}}
\end{itemize}
During evaluation, the \texttt{MMLU-Pro} score is computed as the average over all subsets. For \texttt{AIME2024}, we conduct eight runs and report the mean score. The results of LiveCodeBench come from tests on \texttt{lcb\_code\_generation} and \texttt{lcb\_code\_execution}.

% \begin{figure*}[t]
%   \centering
%   \includegraphics[width=\linewidth]{latex/imgs/motivation_question.pdf}
%   \caption {Classifier-free guidance was applied to all tokens, but only a subset of tokens was successfully guided.}
%   \label{fig:motivation_question}
% \end{figure*}

\subsection{Comparison with SOTA test-time scaling methods}

\begin{table}[h]
\centering
\renewcommand{\arraystretch}{1.15}
\setlength{\tabcolsep}{6pt}
\resizebox{1.0\linewidth}{!}{

\begin{tabular}{ccccccc}
    \hline
    \textbf{Model} & \textbf{Method} & $\mathbf{\tau}$ & \textbf{AIME2024} & \textbf{MATH500} & \textbf{GPQA-D} & \textbf{Avg.} \\
    \hline
    \multirow{8}{*}{\rotatebox{90}{\makebox[0pt][c]{\textbf{Qwen3-8B}}}}
    & \textbf{TALE} & -- & 68.9 & 92.3 & 59.1 & 70.4 \\
    & \textbf{NoThinking} & -- & 30.0 & 87.1 & 54.2 & 57.1 \\
    & \textbf{Dynasor} & -- & 62.2 & 91.7 & 57.7 & 70.5 \\
    & \textbf{DEER} & -- & 45.6 & 88.7 & 59.3 & 64.5 \\
    & \textbf{CGRS} & -- & 61.1 & 93.3 & 59.8 & 71.4 \\
    % & \textbf{VC} & -- & 73.3 & 92.6 & 57.6 & 74.5 \\
    & \multirow{3}{*}{\textbf{Ours}}
    % &   0.1 & 76.6 & 96.8 & 54.6 & 76.0 \\
    & 0.5 & 77.1 & \textbf{97.0} & 60.6 & 78.2 \\
    & & 1.0 & 75.4 & \textbf{97.0} & 60.1 & 77.5 \\
    & & 1.5 & \textbf{78.3} & 95.8 & \textbf{61.1} & \textbf{78.4} \\
    % & & 2.0 & 75.0 & 96.8 & 56.1 & 76.0 \\
    \hline
    \multirow{8}{*}{\rotatebox{90}{\makebox[0pt][c]{\textbf{DeepSeek-R1-7B}}}}
    & \textbf{TALE} & -- & 48.9 & 89.1 & 36.2 & 58.1 \\
    & \textbf{NoThinking} & -- & 32.2 & 80.9 & 37.9 & 50.3 \\
    & \textbf{Dynasor} & -- & 47.8 & 81.8 & 22.2 & 50.6 \\
    & \textbf{DEER} & -- & 47.8 & 89.6 & 33.1 & 56.8 \\
    & \textbf{CGRS} & -- & 52.2 & 87.6 & 32.8 & 57.5 \\
    % & \textbf{VC} & -- & 48.3 & 92.4 & 32.3 & 57.7 \\
    & \multirow{3}{*}{\textbf{Ours}}
    % &   0.1 & 57.1 & 92.8 & 46.0 & 65.3 \\
    & 0.5 & 51.7 & \textbf{93.6} & \textbf{51.52} & 65.6 \\
    & & 1.0 & \textbf{62.9} & 91.8 & 48.5 & \textbf{67.7} \\
    & & 1.5 & 60.0 & 93.2 & 40.9 & 64.7 \\
    % & & 2.0 & 58.8 & 90.6 & 38.9 & 62.8 \\
    \hline
\end{tabular}
}
\caption{Comparison with SOTA test-time scaling methods. Ours (MTI) consistently outperforms SOTA test-time scaling methods.}
\label{tab:comparison_with_SOTA_methods}
\end{table}

As shown in Tab.~\ref{tab:comparison_with_SOTA_methods}, we compare our method with the state-of-the-art approaches TALE~\citep{TALE}, NoThinking~\citep{NoThinking}, Dynasor~\citep{Dynasor}, DEER~\citep{DEER}, and CGRS~\citep{CGRS} under the same settings on Qwen3-8B and DeepSeek-R1-7B across AIME2024, Math500, and GPQA-Diamond. The results demonstrate the superiority of our method.

\begin{table}[h]
\centering
\renewcommand{\arraystretch}{1.15}
\setlength{\tabcolsep}{6pt}
\resizebox{0.7\linewidth}{!}{
% \resizebox{1.0\linewidth}{!}{
\begin{tabular}{ccc}
    \hline
    \textbf{Method} & \textbf{Entropy scope} & \textbf{AIME2024(\%)}\\
    \hline
    \textbf{DI}    & -- & 73.75 \\
    \textbf{VC}    & -- & 73.34  \\
    \multirow{2}{*}{\textbf{Ours}}
        & $Entropy \le 1.5$ & 71.67 \\
        & $Entropy > 1.5$ & 78.34 \\
    \hline
\end{tabular}
}
\caption{Performance under different entropy scopes. The performance gains in high-entropy scopes significantly outweigh those in low-entropy ones.}
\label{tab:energy_scope}
\end{table}

\subsection{Comparison with contrastive decoding}
\begin{table*}[t]
\centering
\small

\begin{adjustbox}{width=1.0\linewidth}
\begin{tabular}{lcccccccccccccccc}
\toprule
Method 
& DI 
& VCD 
& \multicolumn{6}{c}{MTI (CD)} 
& VCD 
& \multicolumn{6}{c}{MTI (CD)}
\\
\midrule
% ---------------- alpha row ------------------
$\alpha$
& -
& 0.5
& \multicolumn{6}{c}{0.5}
& 1.5
& \multicolumn{6}{c}{1.5}
\\[-1mm]

\cmidrule(lr){4-9}
\cmidrule(lr){11-16}

% ---------------- tau row ------------------
$\tau$
& -
& -
& 0.1 & 0.5 & 1.0 & 1.5 & 2.0 & -
& -
& 0.1 & 0.5 & 1.0 & 1.5 & 2.0 & -
\\
\midrule

% ---------------- AIME row ------------------
AIME2024 (\%)
& 60.00 & 65.00
& \textbf{70.00} & 68.34 & 68.34 & 66.67 & 61.67 & -
& 3.33 & 1.67 & 1.67 & 10.00 & 51.67 & 66.67 & -
\\

% ---------------- MTI Usage row ------------------
MTI Usage
& - & 100\%
& \textbf{16.6\%} & 8.2\% & 1.4\% & 0.2\% & 0.1\% & -
& 100\% & 25.9\% & 17.5\% & 11.3\% & 8.8\% & 0.1\% & -
\\

\bottomrule
\end{tabular}
\end{adjustbox}
    \caption{Performance of Ours(MTI) with Contrastive Decoding (CD) on Ling-mini-2.0 using the AIME2024 dataset. VCD denotes the vanilla contrastive decoding baseline. We report results across various hyperparameter configurations for $\alpha$ and $\tau$, along with the corresponding MTI usage percentage. As shown CFG outperforms CD.}
\label{tab:CD}
\end{table*}
Contrastive Decoding (CD)~\citep{CD} is introduced to suppress unreliable generations by contrasting a strong model with a weaker counterpart, thereby reducing hallucinations and improving faithfulness to the conditioning context.
The formulation for CD is defined as:
\begin{equation}
% \begin{aligned}
\begin{array}{l}
\log \hat{P}\left(x_t \mid c,\bar{c}, x_{<t}\right) \\
  = \log P_s \left(x_t \mid c, x_{<t}\right) \\
  \quad {}
  - \alpha \cdot \log P_w \left(x_t \mid \bar{c}, x_{<t}\right),
\end{array}
% \end{aligned}
\label{eq:cd}
\end{equation}
where $\alpha$ is a hyperparameter and $\log p_s$ denotes the distribution from a large expert LM, while $\log p_w$ is derived from a small amateur LM. Notably, if $\log p_w$ is replaced by a version of the expert model itself, the formulation is closely resembles Classifier-Free Guidance (CFG). To evaluate the functional distinctions between these two methods, we integrate CD into the MTI framework—replacing the original CFG component—and perform a comparative analysis to observe their respective effects on generation quality.

We evaluate the performance on the Ling-mini-2.0 model. The results show that $\alpha = 0.5$ yields better performance than $\alpha = 1.5$. The optimal result is achieved when $\tau = 0.1$ and $\alpha = 0.5$; while this configuration outperforms both DI and VCD, it remains below the 71.25\% achieved by the CFG-based version of MTI (see Tab.~\ref{tab:aime2024}). Furthermore, a comparison of inference overhead reveals that MTI(CD) requires the execution of CD on 16.6\% of tokens, whereas MTI(CFG) requires CFG on only 1.4\% of tokens. This proves that the effectiveness of CFG is superior to that of CD.
% 这证明CFG的有效性要更优于CD。

\subsection{Performance under different entropy scopes.}
\label{sec:different_entropy_scopes}

Following the experimental setup in Tab.~\ref{tab:aime2024}, we evaluate Qwen3-8B on the AIME2024 benchmark across various entropy thresholds. The results reveal a clear performance divide: applying CFG to low-entropy tokens ($\text{entropy} \le 1.5$) leads to a significant degradation compared to Direct Inference (DI). This demonstrates that interventions on low-entropy regions are inherently detrimental, as they introduce instability that accumulates over the long-form reasoning chains required by AIME2024. In contrast, by masking out these low-entropy tokens and restricting CFG only to high-entropy regions ($\text{entropy} > 1.5$), MTI significantly outperforms Vanilla CFG (VC). This confirms that selective intervention is crucial; MTI achieves superior results precisely by exempting predictable tokens from guidance, thereby maintaining stability while leveraging the benefits of CFG where it is most effective.

\subsection{Case study.}

\begin{figure}[h]
    \newtcolorbox{scopebox}[2]{%
        left=0pt,
        right=0pt,
        top=0pt,
        bottom=0pt,
        boxsep=3pt,
        colback=white,
        colframe=black,
        title={#1},
    }
    \centering
    \scriptsize
    \begin{scopebox}{Comparing Reasoning Outputs on Math}{}
        \noindent\textbf{Problem.}
        Consider the function
        \[
        f(x)=
        \begin{cases}
        ax^2 & \text{if } x \ge a,\\
        ax + 2a & \text{if } x < a,
        \end{cases}
        \]
        where $a$ is a real number.
        What is the largest value of $a$ such that the graph of $y=f(x)$ intersects every horizontal line at least once?
        Please reason step by step, and put your final answer within ().

        \begin{scopebox}{Direct Inference: Upper bound miscalculated; lost the $+2a$ term.}{}
            \wrong{``... $f(x)=ax+2a,\; x<a$ gives range $(-\infty,a^2)$, hence is (1) ''}
        \end{scopebox}
        
        \begin{scopebox}{Vanilla CFG: Incorrect reasoning; Repeated generation.}{}
            \wrong{``... So for any $a \neq 0$, the function is surjective onto $\mathbb{R}$. \\ But wait... Let me think again... But wait... Let me think again... ''}
        \end{scopebox}

        \begin{scopebox}{Ours: Correct bound and inequality; final answer is correct.}{}
            \correct{``... union is $(-\infty,a^2+2a)\cup[a^3,\infty)$, so the maximum $a$ is (2) ''}
        \end{scopebox}
    \end{scopebox}

    \caption{
        Case study: direct inference vs. vanilla CFG vs. our method on Qwen3-8B.%-Base.
    }
    \label{fig:case study}
\end{figure}

As shown in Fig.~\ref{fig:case study}, we observe two representative failure modes. Direct Inference: symbolic information loss—when rewriting the upper bound of the piecewise range, the constant term ``$+2a$'' is dropped, yielding an underestimated bound and a wrong interval. Vanilla CFG: the main issue is looped generation: the model repeatedly oscillates between claims (e.g., surjectivity onto $\mathbb{R}$) and retractions without a verifiable stopping criterion, and it neglects boundary conditions of the piecewise form. 
This looping arises from applying the CFG to all tokens, which can induce reasoning even for low-entropy tokens, thereby disrupting an otherwise stable reasoning process. Ours: By enforcing explicit boundary/inequality checks and composing the ranges in a unified manner, it avoids these pitfalls and produces the correct final answer. 
% More results can be found in Appendix.~\ref{sec:coding_case_study}.

% From Fig.~\ref{fig:coding_case_study}, we observe that the direct inference method misinterprets the problem and produces an incorrect analysis. The vanilla CFG method exhibits the same issue. Our method introduces perturbations at key nodes of LLM reasoning, stabilizing the reasoning chain and thereby generating the correct result.

\subsection{Discussion on task-specific negative prompts}

Following the same experimental setting as in Tab.~\ref{tab:main_results}, we evaluate task-specific negative prompts using DeepSeek-R1-7B on Math500 and HumanEvalPlus. As shown in Tab.~\ref{tab:task_specific_np}, while \texttt{OUTPUT ERROR} is a robust general-purpose choice, aligning the negative prompt's semantics with domain characteristics can further enhance performance. Specifically, \texttt{Bad Reasoning} is more suitable for mathematical tasks (94.20\% vs. 93.40\% for \texttt{SYNTAX ERROR}) as it better suppresses flawed logical chains. Conversely, \texttt{SYNTAX ERROR} is more effective for coding (77.44\% vs. 76.83\% for \texttt{Bad Reasoning}) by directly targeting structural failure modes. These results demonstrate that performance can be further optimized by refining the semantic granularity of negative prompts to match the specific constraints of the target task.

\label{sec:task_specific_np}
\begin{table}[h]
\centering
\renewcommand{\arraystretch}{1.15} % 稍微增加行间距，更美观
\setlength{\tabcolsep}{6pt}
\resizebox{0.8\linewidth}{!}{
% \resizebox{1.0\linewidth}{!}{
    \begin{tabular}{ccc} % 只有三列：Prompt, Dataset, Accuracy
        \hline
        \textbf{Negative Prompt} & \textbf{Task} & \textbf{Accuracy (\%)} \\
        \hline
        \texttt{Bad Reasoning} & Math & \textbf{94.20} \\
        \texttt{SYNTAX ERROR}  & Math & 93.40 \\
        \texttt{OUTPUT ERROR}  & Math & 93.60 \\
        \hline
        \texttt{Bad Reasoning} & Coding & 76.83 \\
        \texttt{SYNTAX ERROR}  & Coding & \textbf{77.44} \\
        \texttt{OUTPUT ERROR}  & Coding & \textbf{77.44} \\
        \hline
    \end{tabular}
}
\caption{Ablation study of task-specific negative prompts on math and coding benchmarks.} % 修改了标题
\label{tab:task_specific_np}
\end{table}

\subsection{Discussion on decoding strategies}
\label{sec:ds_related_work}
Decoding strategies play a critical role in inference, as they determine how effectively a model’s capabilities are realized. Early methods such as greedy search select the most probable token at each step, but often suffer from repetition. Beam Search mitigates this issue by maintaining multiple high-probability candidates, yet its deterministic nature leads to outputs with limited diversity. To improve diversity, sampling-based approaches have been proposed, including temperature scaling, top-k sampling~\citep{topk}, and top-p sampling~\citep{topp, minp, topn}. These methods increase diversity, but do not correct potential misranking among candidate tokens. MTI targets high-diversity (high-entropy) generation positions during reasoning, where probability ordering is more likely to be unreliable, and applies selective distribution correction to adjust the relative likelihoods of candidate tokens. Accordingly, MTI preserves necessary diversity while mitigating errors caused by miscalibrated probability rankings, and is orthogonal to and compatible with standard sampling strategies.

Other decoding approaches rely on auxiliary models, such as Speculative Decoding~\citep{SD} for acceleration and Contrastive Decoding~\citep{CD, UCD, dola} for reducing hallucinations. In contrast, MTI requires no additional models and focuses on improving reasoning ability rather than speed or hallucination control. LLM-CFG~\citep{cfg_llm} guides generation via unconditional predictions but requires maintaining two KV caches and often yields unstable gains in modern reasoning models. By contrast, MTI employs Selective CFG Intervention only at high-diversity positions, achieving gains with negligible overhead, while Lightweight Negative-Prompt Guidance reuses the conditional KV cache and avoids maintaining dual KV caches, enabling easy integration with modern inference frameworks~\citep{vllm}.

\subsection{Negative prompts in CFG for LLMs}
\label{sec:Negative_prompts_in_CFG_for_LLMs}

\begin{figure}[h]
    \newtcolorbox{scopebox}[2]{%
        left=0pt,
        right=0pt,
        top=0pt,
        bottom=0pt,
        boxsep=3pt,
        colback=white,
        colframe=black,
        title={#1},
    }
    \centering
    \scriptsize
    \begin{scopebox}{Negative prompts in CFG for LLMs}{}
        \noindent\textbf{Prompt: }Describe a dragon in only an extremely concise paragraph.

        \noindent\textbf{Negative prompt: }Describe its flying ability.

        \begin{scopebox}{Without CFG}{}
            A dragon is a mythical creature, often depicted as a large, serpentine being with wings, \textbf{\color{red}capable of flight}, and possessing fire-breathing abilities, symbolizing power, danger, and ancient magic in various cultures.
        \end{scopebox}
        
        \begin{scopebox}{With CFG + Negative prompt}{}
            A dragon is a mythical creature, typically depicted as a large, serpentine being with wings, capable of breathing fire, and often symbolizing power, danger, or wisdom in various cultures.
        \end{scopebox}

    \end{scopebox}

    \caption{
        Illustration of negative prompting in CFG for LLMs. The negative prompt suppresses undesired semantic content from the final generation.
    }
    \label{fig:case_study_cfg_np}
\end{figure}

Classifier-Free Guidance (CFG) originated from diffusion models and is designed to enhance instruction-following capability. In NLP, it can also be applied to improve instruction control in large language models (LLMs). During inference, CFG incorporates both conditional and unconditional predictions. The conditional prediction uses the user prompt as input, while the unconditional prediction uses a negative prompt that specifies undesired content. By linearly combining the two, the final token prediction is guided closer to the conditional prediction and further from the unconditional one, enabling finer control over generation.

For example, as shown in Fig.~\ref{fig:case_study_cfg_np}, the LLM output without CFG contains the phrase ``capable of flight.'' 
By applying CFG with the negative prompt ``Describe its flying ability,''
the reference to flying ability is removed in the final output.

\subsection{Failure cases analysis}
\label{sec:failure_cases_analysis}

\begin{figure}[h]
    \newtcolorbox{scopebox}[2]{%
        left=0pt,
        right=0pt,
        top=0pt,
        bottom=0pt,
        boxsep=3pt,
        colback=white,
        colframe=black,
        title={#1},
    }
    \centering
    \scriptsize

    \begin{scopebox}{Failure cases}{}
        \noindent\textbf{Prompt: }What is the sum of the digits in the terminating decimal representation of the fraction 
        \[
        \frac{4321}{5^7 \cdot 2^8}?
        \]
        Please reason step by step, and put your final answer within $\boxed{}$.

        \noindent\textbf{Gold Answer: }14

        \begin{scopebox}{$MTI(\tau=2.5)$}{}
            We are asked to find the sum of the digits in the terminating decimal representation of the fraction
            \[
            \frac{4321}{5^7 \cdot 2^8}.
            \]

            \noindent{Step 1:} Understand the structure of the denominator. The denominator is
            \[
            5^7 \cdot 2^8.
            \]
            We can write this as
            \[
            \wrong{\underline{2^8 \cdot 5^7 = 2^7 \cdot 2 \cdot 5^7 = 2^7 \cdot 10^7}.}
            \]

            \noindent$\ldots$

            Now sum them:
            \[
            3 + 3 + 7 + 5 + 7 + 8 + 1 + 2 + 5 = 41.
            \]

            Final Answer:
            \[
            \boxed{41}
            \]
        \end{scopebox}

        \begin{scopebox}{$MTI(\tau=2.0)$}{}
            We are asked to find the sum of the digits in the terminating decimal representation of the fraction
            \[
            \frac{4321}{5^7 \cdot 2^8}.
            \]

            \noindent{Step 1:} Understand the \textbf{De(structure)}nominator. The denominator is
            \[
            5^7 \cdot 2^8.
            \]
            We can simplify this by \textbf{combining(nothing)} powers of 2 and 5:
            \[
            \correct{\underline{5^7 \cdot 2^8 = 2^7 \cdot 5^7 \cdot 2^1 = (2 \cdot 5)^7 \cdot 2 = 10^7 \cdot 2}.}
            \]

            \noindent$\ldots$

            Now sum them:
            \[
            0 + 0 + 0 + 2 + 1 + 6 + 0 + 5 = 14.
            \]

            Final Answer:
            \[
            \boxed{14}
            \]
        \end{scopebox}

    \end{scopebox}

    \caption{
        Failure cases analysis.
    }
    \label{fig:mti_math_example}
\end{figure}

When the entropy threshold ($\tau$) is set too high, performance may degrade because interventions skip critical high-entropy tokens. The following failure case, obtained under the Tab.~\ref{tab:main_results} setup in the main text using Qwen3-8B (non-thinking mode) on the OpenCompass Math500 dataset, compares $MTI_{\tau=2.0}$ and $MTI_{\tau=2.5}$. Bold text marks positions where $MTI$ is applied, and tokens in parentheses indicate the original tokens before $MTI$. $MTI_{\tau=2.5}$ yields the incorrect reasoning process $\left(\underline{2^8 \cdot 5^7 = 2^7 \cdot 2 \cdot 5^7 = 2^7 \cdot 10^7}\right)$ because the large $\tau$ prevents intervention on key high-entropy tokens, whereas $MTI_{\tau=2.0}$ successfully intervenes on them and produces the correct answer.

\subsection{Use of LLMs}
We use LLMs to polish the paper, correct some grammatical errors, and make the language more concise and fluent.

\end{document}